\pdfoutput=1

\documentclass[11pt]{article}

\usepackage[final]{acl}
\usepackage{times}
\usepackage{latexsym}
\usepackage{graphicx}
\usepackage{subfigure}
\usepackage[T1]{fontenc}

\usepackage[utf8]{inputenc}

\usepackage{microtype}

\usepackage{inconsolata}

\usepackage{microtype}
\usepackage{hyperref}
\usepackage{url}
\usepackage{booktabs}
\usepackage{subcaption}
\usepackage{graphicx}
\usepackage{amsmath}
\usepackage{amssymb}
\usepackage{multirow}
\usepackage{makecell}
\usepackage{colortbl}
\usepackage{xcolor}
\usepackage{wrapfig}
\usepackage{tcolorbox}
\usepackage{xspace}

\title{\textsc{BanTMedQA}: A Multimodal Benchmark for\\
Open-Ended Bangla Medical Question Answering}

\author{
 \textbf{Rafid Ahmed\textsuperscript{1,*}}\quad
 \textbf{Md.\ Faiyaz Abdullah Sayeedi\textsuperscript{2,*}}\quad
 \textbf{Fariha Anjum Shifa\textsuperscript{3}}
\\[2pt]
 \textbf{Mir Sazzat Hossain\textsuperscript{4, 7}}\quad
 \textbf{Faria Iffat Reshma\textsuperscript{5}}\quad
 \textbf{Mahbub E Sobhani\textsuperscript{2}}
\\[2pt]
 \textbf{Md Fahim\textsuperscript{6,\dag}}\quad
 \textbf{Md Farhad Alam Bhuiyan\textsuperscript{7}}
\\[8pt]
 {\small
 \begin{tabular}[t]{@{}c@{}}
  \textsuperscript{1}University of Central Florida, USA \qquad
  \textsuperscript{2}BRAC University, Bangladesh \qquad
  \textsuperscript{3}Virginia Tech, USA \\[1pt]
  \textsuperscript{4}Center for Computational \& Data Sciences, Independent University, Bangladesh \qquad
  \textsuperscript{5}University of Dhaka, Bangladesh \\[1pt]
  \textsuperscript{6}The University of Texas at Dallas, USA \qquad
  \textsuperscript{7}Penta Global Limited, Bangladesh
 \end{tabular}}
\\[6pt]
 {\footnotesize
 \begin{tabular}[t]{@{}c@{}}
  \textsuperscript{*}Equal contribution \qquad
  \textsuperscript{\dag}Project lead \\[1pt]
  \textbf{Correspondence:}
  \texttt{\{ahmedrafid023, fahimcse381\}@gmail.com, msayeedi212049@bscse.uiu.ac.bd}
 \end{tabular}}
}

\newcommand{\name}{\textsc{BanTMedQA}\xspace}

\begin{document}
\maketitle
\begin{abstract}
Large Language Models (LLMs) now perform strongly on medical licensing exams, but solving multiple-choice questions is very different from helping real patients. This is especially important in low-income countries like Bangladesh where doctor shortages and long waiting times make healthcare difficult to access. AI assistants could help by guiding patients before appointments, organizing symptom details, and answering health queries. However, most medical AI benchmarks still rely on textbook-style multiple-choice datasets, which do not reflect real-world scenarios in which patients describe symptoms informally. In Bangladesh, people often communicate using a mix of Bangla, English, and transliterated Bangla, frequently alongside medical images such as X-rays or reports. To address this gap, we introduce the first open-ended, multimodal medical QA dataset for the Bangla language ecosystem. Our dataset includes Standard Bangla, transliterated Bangla, English and image-based patient scenarios to better reflect real-world communication. Using this benchmark, we evaluate leading proprietary and open-source LLMs and identify where current models succeed and fail in realistic low-resource healthcare settings. The dataset is available under gated access on Hugging Face Hub\footnote{\url{https://huggingface.co/datasets/FaiyazAbdullah114708/BanTMedQA}}

\end{abstract}

\section{Introduction}

\begin{figure*}[t]
    \centering
    \includegraphics[width=\textwidth]{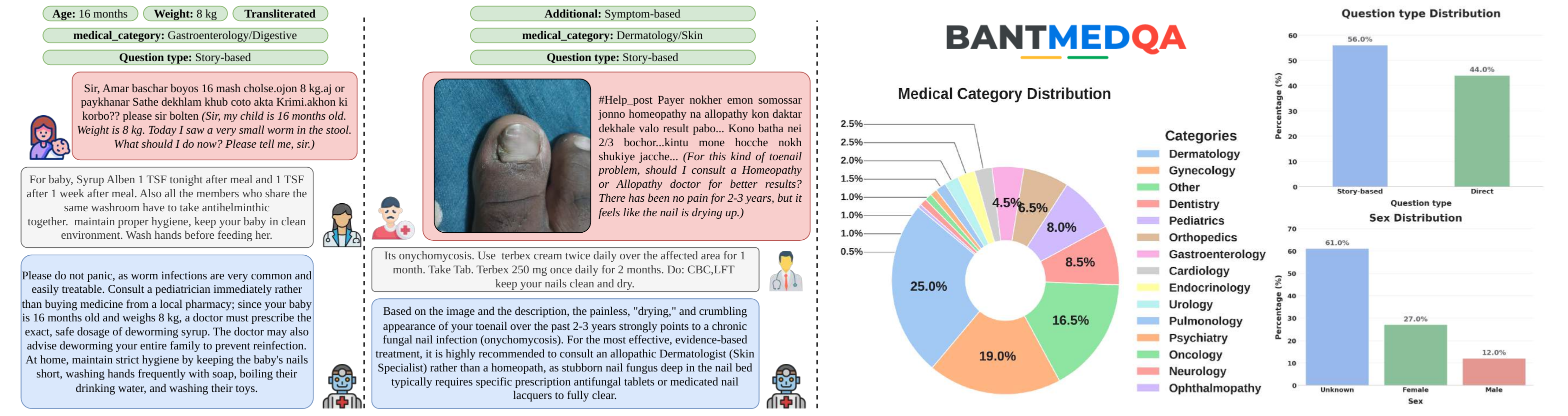}
    \caption{\textbf{Overview of the \name{} Dataset.} \textbf{(Left)} A text-only pediatric query featuring metadata, transliterated Bangla text with English translation, expert physician answer, and AI-generated response. \textbf{(Center)} A multimodal dermatology query paired with a clinical image. \textbf{(Right)} Dataset distributions across medical categories, question types, and patient sex.}
    \label{fig:method}
\end{figure*}

Recent advances in Large Language Models (LLMs) have significantly transformed natural language processing, with state-of-the-art (SOTA) systems demonstrating strong performance across tasks such as question answering, summarization, reasoning, and clinical decision support \cite{brown2020language, singhal2023large}. In healthcare, these developments have accelerated the emergence of AI-driven medical Question Answering (QA) systems capable of interpreting complex medical terminology, supporting clinical workflows, and achieving competitive performance on standardized medical examinations \cite{nori2023capabilities, wang2026capabilities, milad2024assessing, pal2022medmcqa}. As a result, LLMs are increasingly being explored as cognitive assistants that can improve access to medical information, reduce documentation burden, and support healthcare delivery at scale \cite{wang2025trustworthy}.

The potential impact of such systems is particularly important in low and middle income countries (LMICs) like Bangladesh, where healthcare infrastructures face persistent resource constraints \cite{al2025willingness}. A severe shortage of registered physicians, combined with high patient volumes in public hospitals and primary care facilities, creates significant barriers to timely medical consultation \cite{shawon2026using}. Patients often experience long waiting periods before obtaining appointments, while the limited consultation time available during clinical visits can restrict effective communication of symptoms and medical history. These structural challenges frequently reduce the quality and efficiency of patient-centered care.

Conversational AI systems offer a promising way to alleviate some of these pressures. As a pre-consultation support tool, an AI assistant can help patients organize symptoms, describe medical concerns more clearly and better prepare for physician interactions. For non-critical conditions, such systems may also provide preliminary guidance and reduce unnecessary hospital visits, potentially lowering both patient costs and clinical overcrowding \cite{al2025willingness}. However, despite the rapid progress of modern LLMs, an important question remains unanswered: \textit{are current SOTA models sufficiently reliable for deployment in realistic, localized healthcare settings?}

Existing medical NLP evaluation frameworks suggest a substantial gap between benchmark performance and real-world clinical utility. Most current medical QA benchmarks rely heavily on clean, multiple-choice question (MCQ) formats derived from professional examinations \cite{pal2022medmcqa, sultana2025banglamedqa}. While these datasets are useful for measuring factual recall, recent studies show that strong MCQ performance often reflects statistical pattern recognition and test-taking heuristics rather than genuine clinical reasoning \cite{griot-etal-2025-pattern}. In practice, real-world patient interactions are open-ended, conversational and context-dependent, requiring models to interpret detailed symptom histories, ambiguous narratives and temporally evolving clinical conditions \cite{kweon2024ehrnoteqa}. Despite growing interest in scenario-based medical evaluation, no prior work has introduced a dedicated Bangla benchmark designed specifically for open-ended, clinically realistic QA assessment.

This challenge becomes even more complex in Bangladesh’s multilingual digital environment. Everyday online communication frequently involves mixed linguistic forms, including native Bangla script, English and transliterated Bangla written in Roman characters (commonly known as Banglish), often within the same sentence or conversation \cite{fahim-etal-2024-banglatlit, jawad2025benchmarking, jawad20265}. Furthermore, medical consultations are inherently multimodal; patients regularly supplement textual descriptions with images of symptoms, prescriptions, diagnostic reports, or skin conditions. Consequently, clinically useful AI systems must be capable of understanding noisy, script-mixed, and multimodal inputs rather than relying solely on clean textual benchmarks.

To address these limitations, we introduce \name{}, the first open-ended multimodal benchmark dataset designed for Bangla medical question answering in realistic healthcare settings. The dataset contains real-world medical queries collected from a large online medical community in Bangladesh, including both text-only and image-text scenarios. To this end, our contributions are:

\begin{itemize}
\item We introduce \name{}, the first multimodal benchmark dataset for scenario based Bangla medical question answering, designed to reflect realistic healthcare interactions in low-resource settings, covering both text-only and image-text scenarios. 

\item We evaluate SOTA proprietary and open-source LLMs across multilingual and multimodal medical QA settings, analyzing performance across transliterated Bangla, Standard Bangla, English, and image-based clinical scenarios.

\item We identify critical failure modes and limitations of current multilingual and multimodal LLMs in low-resource clinical environments, providing insights into the challenges of deploying AI-driven healthcare systems safely and reliably in real-world settings.

\end{itemize}
\section{Related Work} \label{sec:related_work}

\noindent \textbf{LLMs in Medical QA \& Scenario based Medical QA.} The performance of LLMs is carefully investigated in Medical domain considering different aspects. \citet{pal2022medmcqa} proposed MedMCQA that contains over 194,000 medical MCQs. As models began achieving high scores on these exams, researchers questioned whether these results represented genuine understanding. \citet{griot-etal-2025-pattern} highlighted this issue using a fictional medical benchmark built around an imaginary organ called the Glianorex. Despite having no prior knowledge, LLMs still performed strongly by relying on linguistic patterns and test-taking shortcuts rather than real reasoning. To improve explainability, \citet{kim-etal-2024-medexqa} introduced MedExQA, which includes expert-written rationales for each answer. Their findings showed that generative responses align more closely with physician judgment than simple answer selection. \citet{yang2024textbook} proposed KnoBo, a framework that integrates explicit medical knowledge into the reasoning process. Recent benchmarks increasingly emphasize open-ended, scenario-based evaluation. K-QA \cite{manes2024k} uses real patient-doctor interactions containing ambiguous symptoms and conversational language, while EHRNoteQA \cite{kweon2024ehrnoteqa} evaluates reasoning across longitudinal patient records and multiple hospital visits. These datasets better reflect real-world healthcare settings by requiring contextual, multi-step reasoning rather than simple pattern matching. Overall, the literature shows a clear shift away from multiple-choice evaluation toward open-ended, scenario-driven clinical reasoning. Our work follows this direction by extending scenario-based medical QA into a low-resource linguistic setting.

\medskip

\noindent \textbf{Medical QA in Bangla Language.} Developing automated Question Answering (QA) systems in Bangla is heavily bottlenecked by a scarcity of structured clinical datasets. Early text-based pipelines focused on framework foundations and similarity matching. \citet{zannat2025banglamed} developed BanglaMed-QA, curating a structured medical knowledge base of 4,493 QA pairs across 506 diseases. Their pipeline resolved clinical pronouns via Part-of-Speech tagging and applied a multi-metric similarity voting ensemble to match user queries. \citet{sultana2025banglamedqa} introduced BanglaMedQA and BanglaMMedBench, a benchmark of 2,000 medical multiple-choice questions (MCQs), utilizing an \texttt{Agentic RAG} pipeline over digitized local textbooks. Shifting toward diagnostic imagery, \citet{ahmed2026good} introduced BanglaMedVQA, a multimodal dataset containing 2,000 image-question pairs. Their evaluation revealed that while LVLMs handle generalized visual tasks well, they struggle significantly with specialized diagnostic reasoning. \citet{ahmed2026evaluating} introduced a 7000 image-question pair dataset and fine-tuned the SOTA models using it. Though incorporating LoRA fine-tuning provides substantial improvements, enabling models to surpass closed-source baselines, the limitations
in specialized medical reasoning still persist. 

Existing studies have built important foundations for Bangla medical AI, but they mostly focus on single-turn QA datasets, MCQs, or simple image-based queries. None address open-ended question answering based on detailed patient histories or complex clinical scenarios. This remains a major gap, since handling nuanced, conversational patient narratives is essential for real-world healthcare applications.

\section{\name: Dataset Creation} \label{sec:dataset}
We develop \name dataset to assess LLMs' behavior on authentic transliterated Bangla, standard Bangla and English medical questions across various healthcare domains. This section describes the dataset construction pipeline, which comprises five stages: question collection and filtering, anonymization, data categorization, answer generation by medical experts, and data validation.

\paragraph{Question Collection and Filtering.} \label{para:data_collection}
We sourced a mixture of Bangla, English and transliterated Bangla medical
questions from publicly accessible Facebook groups in which members of the
public in Bangladesh seek advice on medical concerns. Authors browsed these
groups as ordinary users and copied each selected post by hand. No scrapers,
crawlers, bots, browser automation, or platform APIs were used at any stage,
and no access controls were circumvented; we collected only content visible
to any visitor to these groups.
This manual process is consistent with Meta's Terms of Service, which
prohibit accessing or collecting data from Meta's products by automated
means without prior permission (Section~3.2.3).\footnote{Meta Terms of
Service: \url{https://www.facebook.com/terms.php}}
We deliberately do not name the source groups, since combined with the
query text, group names would narrow the search space for locating original
posts. The released dataset contains no Facebook account data, and we claim
no ownership of the original posts.
The initial dataset comprised 3,500 queries, from which we removed any query
containing Personally Identifiable Information (PII) such as a name or
identification number. Additionally, within the image data, any image
revealing user identity of any form was blurred or removed. After filtering, the final
dataset contained 2,000 medical queries.

\paragraph{Anonymization of Queries and Images.} To ensure the privacy of the original posters, we sanitized every retained query and its associated media in three ways (Figure~\ref{fig:anonymisation}). First, we deleted any remaining names of people, nicknames, usernames, @mentions, hashtags, and place names smaller than a city or district, and replaced the names of clinics and diagnostic centres with generic terms (e.g., a named dental clinic became ``a dental clinic''). For dates, we removed the year but kept the day and month, so that time intervals did not change. Second, we lightly paraphrased each query by reordering clauses, substituting near-synonyms, varying spellings, and merging or splitting sentences, so that searching the query text does not lead back to the original post (Appendix~\ref{app:release}). We kept all clinical content unchanged, including symptoms, body parts, durations, age, sex, drug names, doses, test values, and the question being asked, and preserved each query's script and informal, code-mixed style. For each transliterated query, the same edits were applied to its Standard Bangla and English reformulations (Appendix~\ref{app:reformulation}), so these versions remain parallel. Third, for queries accompanied by images, we blurred all surrounding visual details that could potentially reveal personally identifiable information (PII), leaving only the relevant diagnostic region visible.

\paragraph{Data Categorization.}\label{para:data_categorize}
We organized the collected queries into categories based on modality (text / text and image-based) and medical focus, so that the dataset could be used to evaluate LLMs' performance across various medical categories. We also labelled each question according to one of two query types: direct or story-based. Direct queries pose a question without providing patient background, whereas story-based queries first present the patient's history before posing the question. The final dataset consisted of 15 medical categories.
Both labels were assigned by three instruction-tuned LLMs (Claude-Sonnet-4.6, GPT-5-Mini and Gemini-3-Flash), each prompted independently with the taxonomy definitions, and the final label was decided by majority vote. For query type, the three models agreed unanimously on 94.2\% of queries (Fleiss' $\kappa = 0.92$). For medical category, unanimous agreement was 83.4\% ($\kappa = 0.86$); the 32 queries (1.8\%) on which all three models disagreed were labelled by an MBBS-qualified clinician.

\paragraph{Answer Generation by Medical Experts.}\label{para:ans_generation}
The answers to the medical queries were collected by annotators who were 10 MBBS doctors working as general practitioners. Each had completed an MBBS (Bachelor of Medicine, Bachelor of Surgery) degree from a medical college in Bangladesh, followed by a one-year internship. To distribute the queries, each annotator was first asked, via a Google Form, which medical domain they preferred to answer questions in, and queries were assigned based on their stated preferences. Queries were restricted to the scope of general practice, i.e., questions that an MBBS-qualified general practitioner could answer with full confidence; any query requiring specialist expertise beyond this level was omitted from the dataset. For the annotation process, we did not adopt any specific annotation framework, as the annotators were already proficient with Google Drive, Docs, and Sheets. Each annotator's answers were not visible to any other annotator at any point during the annotation process, ensuring independent answer generation. The annotation guidelines provided to the annotators is detailed in Appendix \ref{appendix:guidelines}. The answers were provided in free-text format in a shared Google Sheet. On average, annotators were compensated above the hourly equivalent of
entry-level government physician pay in Bangladesh
(Appendix~\ref{app:compensation}).

\paragraph{Data Validation.}\label{para:validation}
To assess the clinical quality of the collected answers, every query--answer pair in the final dataset was independently rated by two annotators other than the pair's original author, using a two-point scale: 0 (\textit{disagree}, i.e.\ the answer is incorrect, irrelevant, or clinically incomplete) and 1 (\textit{agree}, i.e.\ the answer is medically correct and sufficiently complete to address the query). Ratings reflect correctness and completeness only; an answer was not penalized for being phrased differently than a rater would have phrased it themselves, so long as it remained medically accurate and complete. Raters were blind to the identity of the original annotator, and item order was randomized independently for each rater to avoid ordering effects. A query--answer pair was retained in the final dataset only if both raters agreed (rating 1). Pairs where the two raters disagreed, or where both rated the answer as inadequate, were discarded rather than adjudicated, ensuring every retained answer reflects unanimous clinical endorsement. Under this protocol 90.2\% of pairs were retained; of the 9.8\% discarded, 192 pairs reflected disagreement and 4 were rejected by both raters, giving an observed agreement of 90.4\%. Because 95.0\% of ratings were ``agree'', Cohen's $\kappa$ is uninformative under this prevalence skew \cite{feinstein1990high}; we report $\kappa = -0.011$ alongside Gwet's AC$_1 = 0.894$, which is stable under skew (Appendix~\ref{app:agreement}). This yielded the final dataset of 1{,}804 examples.

\paragraph{Dataset statistics.}
After filtering, the final dataset comprises 1,804 medical queries paired with expert-generated answers. By query type, 56.0\% are story-based, while the remaining 44.0\% are direct queries. In terms of patient sex, the majority of queries do not specify this attribute (61.0\%), whereas 27.0\% concern female and 12.0\% male patients. The dataset spans 15 medical categories (Figure~\ref{fig:method}, right), with the largest proportions in Dermatology (25.0\%) and Gynecology (19.0\%), followed by Dentistry (8.5\%), Pediatrics (8.0\%) and Orthopedics (6.5\%). The remaining categories including Gastroenterology, Cardiology, Endocrinology, Urology, Pulmonology, Psychiatry, Oncology, Neurology and Ophthalmology each constitute 4.5\% or less and 16.5\% of queries are grouped under ``Other". The ``Other" category consists of general medicine questions, multi-systemic complaints like fever, a small number of hematology queries and generalized health advice that fall outside the scope of the predefined specialist domains. On average, input queries contain 33.4 words (189.8 characters), while the corresponding expert answers contain 31.8 words (181.6 characters).
\begin{table*}[ht]
\centering
\small
\setlength{\tabcolsep}{4pt}
\begin{tabular}{l ccc ccc ccc}
\toprule
\multirow{2}{*}{\textbf{Model}}
  & \multicolumn{3}{c}{\textbf{Transliterated Bangla}}
  & \multicolumn{3}{c}{\textbf{Standard Bangla}}
  & \multicolumn{3}{c}{\textbf{English}} \\
\cmidrule(lr){2-4} \cmidrule(lr){5-7} \cmidrule(lr){8-10}
 & \textbf{BScore} & \textbf{LAVE} & \textbf{Human}
 & \textbf{BScore} & \textbf{LAVE} & \textbf{Human}
 & \textbf{BScore} & \textbf{LAVE} & \textbf{Human} \\
\midrule
\multicolumn{10}{l}{\textit{\textbf{Closed-Source Models}}} \\
\quad Claude-Sonnet-4.6
  & \cellcolor{cyan!10}\textbf{58.23} & \cellcolor{cyan!10}\textbf{69.60} & \cellcolor{cyan!10}\textbf{57.53}
  & 43.56 & \cellcolor{cyan!10}\textbf{82.50} & 65.70
  & \cellcolor{cyan!10}\textbf{46.82} & 78.33 & 67.57 \\
\quad Gemini-3-Flash
  & 58.18 & 64.78 & 48.15
  & 40.36 & 75.83 & \cellcolor{cyan!10}\textbf{70.34}
  & 46.13 & 71.67 & \cellcolor{cyan!10}\textbf{73.77} \\
\quad GPT-5-Mini
  & 51.77 & 64.93 & 35.39
  & 43.80 & 80.17 & 57.60
  & 46.31 & \cellcolor{cyan!10}\textbf{80.83} & 60.91 \\
\midrule
\multicolumn{10}{l}{\textit{\textbf{Open-Source Models}}} \\
\quad Qwen3-VL-8B
  & 49.77 & 48.87 & 29.09
  & 42.83 & 59.17 & 41.57
  & 43.80 & 60.00 & 55.51 \\
\quad Gemma-4-E4B-it
  & 54.62 & 57.55 & 42.30
  & \cellcolor{cyan!10}\textbf{46.10} & 70.83 & 58.25
  & 46.65 & 61.67 & 50.03 \\
\quad Llama-3.2-11B-V
  & 41.31 & 29.53 & 16.67
  & 35.14 & 38.33 & 32.18
  & 43.14 & 54.17 & 47.91 \\
\midrule
\multicolumn{10}{l}{\textit{\textbf{Medical LLMs}}} \\
\quad MedGemma-4B
  & 45.08 & 29.64 & 20.60
  & 40.39 & 55.83 & 53.25
  & 43.32 & 57.50 & 55.50 \\
\quad LLaVA-Med-v1.5-Mistral-7B
  & 7.34  & 2.55  & 1.75
  & 19.00 & 2.85  & 0.75
  & 23.69 & 18.75 & 5.75 \\
\bottomrule
\end{tabular}
\caption{Medical VQA evaluation results across different question formulations using BERTScore (BScore, \%), LAVE (\%), and human evaluation scores. Bold values with cyan shading indicate the top score in each metric--language column.}
\label{tab:medical_vqa_results}
\end{table*}

\begin{table*}[htbp]
\centering
\small
\setlength{\tabcolsep}{4pt}
\begin{tabular}{l cc cc cc}
\toprule
\multirow{2}{*}{\textbf{Model}}
  & \multicolumn{2}{c}{\textbf{Transliterated Bangla}}
  & \multicolumn{2}{c}{\textbf{Standard Bangla}}
  & \multicolumn{2}{c}{\textbf{English}} \\
\cmidrule(lr){2-3} \cmidrule(lr){4-5} \cmidrule(lr){6-7}
 & \textbf{Multimodal} & \textbf{Unimodal}
 & \textbf{Multimodal} & \textbf{Unimodal}
 & \textbf{Multimodal} & \textbf{Unimodal} \\
\midrule
\multicolumn{7}{l}{\textit{\textbf{Closed-Source Models}}} \\
\quad Claude-Sonnet-4.6 & \cellcolor{cyan!10}\textbf{59.21} & \cellcolor{cyan!10}\textbf{70.74} & 85.00 & \cellcolor{cyan!10}\textbf{81.25} & 60.00 & 82.00 \\
\quad Gemini-3-Flash    & 51.32 & 60.80 & 82.50 & 72.50 & \cellcolor{cyan!10}\textbf{75.00} & 71.00 \\
\quad GPT-5-Mini        & 46.05 & 60.80 & \cellcolor{cyan!10}\textbf{88.89} & 76.25 & 70.00 & \cellcolor{cyan!10}\textbf{83.00} \\
\midrule
\multicolumn{7}{l}{\textit{\textbf{Open-Source Models}}} \\
\quad Qwen3-VL-8B       & 38.82 & 38.51 & 55.00 & 61.25 & 60.00 & 60.00 \\
\quad Gemma-4-E4B-it    & 41.45 & 50.00 & 72.50 & 70.00 & 55.00 & 63.00 \\
\quad Llama-3.2-11B-V   & 15.79 & 24.72 & 45.00 & 35.00 & 50.00 & 55.00 \\
\midrule
\multicolumn{7}{l}{\textit{\textbf{Medical LLMs}}} \\
\quad MedGemma-4B       & 29.61 & 25.00 & 57.50 & 55.00 & 35.00 & 62.00 \\
\quad LLaVA-Med-v1.5-Mistral-7B & 2.80 & 2.20 & 3.00 & 2.70 & 37.50 & 16.07 \\
\bottomrule
\end{tabular}
\caption{LAVE score (\%) on the image-bearing (Multimodal) and text-only (Unimodal) partitions. The partitions contain different queries, so this compares partition difficulty, not the contribution of the image (see Table~\ref{tab:paired-image}). Bold values with cyan shading indicate the top score in each column.}
\label{tab:unimodal-multimodal}
\end{table*}

\begin{table*}[htbp]
\centering
\small
\setlength{\tabcolsep}{4pt}
\begin{tabular}{l cc cc cc}
\toprule
\multirow{2}{*}{\textbf{Model}}
  & \multicolumn{2}{c}{\textbf{Transliterated Bangla}}
  & \multicolumn{2}{c}{\textbf{Standard Bangla}}
  & \multicolumn{2}{c}{\textbf{English}} \\
\cmidrule(lr){2-3} \cmidrule(lr){4-5} \cmidrule(lr){6-7}
 & \textbf{w/ Image} & \textbf{w/o Image}
 & \textbf{w/ Image} & \textbf{w/o Image}
 & \textbf{w/ Image} & \textbf{w/o Image} \\
\midrule
\multicolumn{7}{l}{\textit{\textbf{Closed-Source Models}}} \\
\quad Claude-Sonnet-4.6 & \cellcolor{cyan!10}\textbf{59.21} & \cellcolor{cyan!10}\textbf{34.21} & \cellcolor{cyan!10}\textbf{61.18} & \cellcolor{cyan!10}\textbf{42.11} & \cellcolor{cyan!10}\textbf{63.16} & \cellcolor{cyan!10}\textbf{44.74} \\
\quad Gemini-3-Flash    & 51.32 & 31.58 & 53.29 & 34.21 & 54.61 & 33.55 \\
\quad GPT-5-Mini        & 46.05 & 11.18 & 51.32 & 17.11 & 60.53 & 13.16 \\
\midrule
\multicolumn{7}{l}{\textit{\textbf{Open-Source Models}}} \\
\quad Qwen3-VL-8B       & 39.47 & 19.74 & 38.16 & 28.29 & 43.42 & 33.55 \\
\quad Gemma-4-E4B-it    & 41.45 & 29.61 & 40.13 & 31.58 & 42.76 & 31.58 \\
\quad Llama-3.2-11B-V   & 33.55 & 8.55  & 21.05 & 21.71 & 38.82 & 43.42 \\
\midrule
\multicolumn{7}{l}{\textit{\textbf{Medical LLMs}}} \\
\quad MedGemma-4B       & 23.03 & 15.79 & 33.55 & 25.66 & 33.55 & 30.26 \\
\quad LLaVA-Med-v1.5-Mistral-7B & 5.88 & 7.35 & 13.57 & 6.43 & 33.82 & 31.62 \\
\bottomrule
\end{tabular}
\caption{Paired image ablation (LAVE, \%). Each image-bearing query is evaluated with its image (w/ Image) and with the identical text alone (w/o Image), so the difference isolates the contribution of the image. Bold values with cyan shading indicate the top score in each column.}
\label{tab:paired-image}
\end{table*}

\section{Experimental Setup}
\label{sec:experiment_setup}

\paragraph{Models.}
We benchmark eight vision-language models (VLMs) spanning three categories: frontier closed-source, open-source general-purpose, and medical-specialized VLMs. For closed-source systems, we evaluate \texttt{Claude-Sonnet-4.6}, \texttt{Gemini-3-Flash}, and \texttt{GPT-5-Mini} through the OpenRouter API, treating them as high-capacity proprietary reference models for multilingual clinical reasoning. For open-source general-purpose VLMs, we evaluate \texttt{Qwen3-VL-8B-Instruct}, \texttt{Gemma-4-E4B-it}, and \texttt{Llama-3.2-11B-Vision-Instruct}, which cover major open-source VLM families and fall within a parameter range suitable for practical self-hosted deployment in resource-constrained healthcare settings, where proprietary APIs may be limited by cost or privacy constraints. Finally, we include two medical-specialized models, \texttt{MedGemma-4B-it} and \texttt{LLaVA-Med-v1.5-Mistral-7B}, both of which incorporate medical-domain adaptation through instruction tuning on clinical or biomedical data.

\paragraph{Inference protocol.}
We use a unified system prompt instructing each model to generate a concise, evidence-based answer strictly in English, regardless of the input language or script. This constraint ensures that outputs from the Transliterated Bangla, Standard Bangla, and English conditions are evaluated against the same expert-written English reference answer, thereby isolating the effect of input representation from potential confounds introduced by multilingual output generation. We use greedy decoding with temperature set to 0, \texttt{do\_sample=false}, and a maximum generation length of 512 tokens. The complete prompt template is provided in Appendix~\ref{all_prompts}. Closed-source models are queried via the OpenRouter API, while open-source models are served locally using the HuggingFace Transformers library on a single NVIDIA L40S GPU with 48\,GB VRAM.

\paragraph{Evaluation metrics.}
We evaluate model responses using three complementary signals that capture semantic similarity, answer correctness, and clinical adequacy. First, \textit{BERTScore (BScore)}~\citep{zhang2019bertscore} measures token-level semantic similarity between the generated response and the expert reference answer using contextual embeddings from DeBERTa-xlarge-mnli, providing a reference-based automatic score that is more robust to paraphrasing than exact lexical overlap, although still limited in assessing factual and clinical correctness. Second, LAVE (LLM-Assisted VQA Evaluation)~\cite{manas2024improving} uses an LLM judge to rate how well a
candidate answer agrees with the expert reference answer. Because our references are physician-written gold answers that already encode the correct response to the query and image, we score each candidate against the
reference rather than re-presenting the question, keeping the evaluation anchored to the clinical gold standard. We have used \texttt{GPT-4.1 Mini} to calculate the LAVE score. Unlike embedding-based similarity, LAVE tolerates clinically valid responses that differ lexically from the reference and we report it as a secondary automatic signal alongside BScore. Finally, \textit{human evaluation} serves as our primary evaluation protocol: MBBS-qualified medical practitioners assess each response against the expert reference answer using the same five-point correctness scale used for LAVE (Appendix~\ref{app:human_eval}), ranging from 0.0 (\textit{completely incorrect}) to 1.0 (\textit{completely correct}), with intermediate levels capturing partially correct or mostly correct responses. This shared scale allows human and LLM-judge scores to be directly compared under an identical rating rubric. Two physicians independently rated all 600 responses in the evaluation subset (quadratic-weighted $\kappa_w = 0.76$; Appendix~\ref{app:human_eval}). All scores are normalized to percentages for reporting, and we prioritize human evaluation because automated metrics often fail to capture clinically important omissions, hallucinations, and reasoning errors in medical generation tasks~\citep{zheng2023judging,wang-etal-2024-large-language-models-fair}.

\section{Results and Analysis} \label{sec:result}

\begin{table*}[ht]
\centering
\setlength{\tabcolsep}{4.0pt}
\footnotesize
\begin{tabular}{l ccc ccc ccc}
\toprule
\multirow{2}{*}{\textbf{Category}}
  & \multicolumn{3}{c}{\textbf{Claude-Sonnet-4.6}}
  & \multicolumn{3}{c}{\textbf{Gemini-3-Flash}}
  & \multicolumn{3}{c}{\textbf{GPT-5-Mini}} \\
\cmidrule(lr){2-4}\cmidrule(lr){5-7}\cmidrule(lr){8-10}
 & \textbf{T-BN} & \textbf{S-BN} & \textbf{EN}
 & \textbf{T-BN} & \textbf{S-BN} & \textbf{EN}
 & \textbf{T-BN} & \textbf{S-BN} & \textbf{EN} \\
\midrule
Cardiology        & 83.33 & 91.67 & 83.33 & 75.00 & 91.67 & 75.00 & 83.33 & \cellcolor{cyan!10}\textbf{100.00} & 83.33 \\
Dentistry/Oral    & 68.75 & \cellcolor{cyan!10}\textbf{87.50} & 81.25 & 50.00 & 62.50 & 56.25 & 75.00 & 81.25 & \cellcolor{cyan!10}\textbf{87.50} \\
Dermatology       & 53.85 & \cellcolor{cyan!10}\textbf{64.42} & 62.50 & 50.00 & 63.46 & 56.73 & 42.31 & 55.77 & 61.54 \\
Endocrinology     & \cellcolor{cyan!10}\textbf{100.00} & \cellcolor{cyan!10}\textbf{100.00} & 93.75 & 91.67 & 93.75 & 95.83 & 91.67 & 95.83 & 95.83 \\
Gastroenterol.    & 60.71 & \cellcolor{cyan!10}\textbf{92.86} & 89.29 & 53.57 & 71.43 & 67.86 & 50.00 & 75.00 & 82.14 \\
Gynecol./OB       & 77.00 & \cellcolor{cyan!10}\textbf{95.00} & 88.00 & 65.00 & 83.00 & 74.00 & 65.00 & 85.00 & 82.00 \\
Neurology         & 62.50 & 75.00 & 68.75 & 37.50 & 43.75 & 50.00 & 75.00 & \cellcolor{cyan!10}\textbf{87.50} & 81.25 \\
Oncology          & 25.00 & 33.33 & 33.33 & \cellcolor{cyan!10}\textbf{100.00} & \cellcolor{cyan!10}\textbf{100.00} & \cellcolor{cyan!10}\textbf{100.00} & \cellcolor{cyan!10}\textbf{100.00} & \cellcolor{cyan!10}\textbf{100.00} & \cellcolor{cyan!10}\textbf{100.00} \\
Ophthalmology     & \cellcolor{cyan!10}\textbf{100.00} & \cellcolor{cyan!10}\textbf{100.00} & \cellcolor{cyan!10}\textbf{100.00} & 75.00 & 83.33 & 83.33 & 75.00 & 83.33 & 83.33 \\
Orthopedics       & 55.56 & \cellcolor{cyan!10}\textbf{77.78} & 72.22 & 52.78 & 66.67 & 61.11 & 50.00 & 66.67 & 69.44 \\
Pediatrics        & 58.33 & \cellcolor{cyan!10}\textbf{66.67} & 62.50 & 45.83 & 50.00 & 56.25 & 37.50 & 64.58 & 64.58 \\
Psychiatry        & 75.00 & \cellcolor{cyan!10}\textbf{83.33} & 75.00 & 75.00 & \cellcolor{cyan!10}\textbf{83.33} & \cellcolor{cyan!10}\textbf{83.33} & 75.00 & \cellcolor{cyan!10}\textbf{83.33} & 75.00 \\
Pulmonology       & 75.00 & \cellcolor{cyan!10}\textbf{91.67} & \cellcolor{cyan!10}\textbf{91.67} & 66.67 & 83.33 & 75.00 & 41.67 & 66.67 & 83.33 \\
Urology           & 75.00 & \cellcolor{cyan!10}\textbf{91.67} & \cellcolor{cyan!10}\textbf{91.67} & 75.00 & 83.33 & 75.00 & 50.00 & 75.00 & 83.33 \\
Other             & 74.04 & \cellcolor{cyan!10}\textbf{86.54} & 81.73 & 58.65 & 77.88 & 65.38 & 62.50 & 82.69 & 79.81 \\
\midrule
\textbf{Average}  & 69.60 & \cellcolor{cyan!10}\textbf{82.50} & 78.33 & 64.78 & 75.83 & 71.67 & 64.93 & 80.17 & 80.83 \\
\bottomrule
\end{tabular}
\caption{Fine-grained performance breakdown (Avg.\ LAVE Score) across medical specialties for closed-source models. The evaluation highlights the performance of Transliterated Bangla (\textbf{T-BN}), Standard Bangla (\textbf{S-BN}) and English (\textbf{EN}) queries.}
\label{tab:lave_proprietary_models}
\end{table*}

\begin{table*}[ht]
\centering
\setlength{\tabcolsep}{4.0pt}
\scriptsize
\begin{tabular}{l ccc ccc ccc ccc ccc}
\toprule
\multirow{2}{*}{\textbf{Category}}
  & \multicolumn{3}{c}{\textbf{Qwen3-VL-8B}}
  & \multicolumn{3}{c}{\textbf{Gemma-4-E4B-it}}
  & \multicolumn{3}{c}{\textbf{Llama-3.2-11B-V}}
  & \multicolumn{3}{c}{\textbf{MedGemma-4B}}
  & \multicolumn{3}{c}{\textbf{LLaVA-Med-v1.5-Mistral-7B}} \\
\cmidrule(lr){2-4}\cmidrule(lr){5-7}\cmidrule(lr){8-10}\cmidrule(lr){11-13}\cmidrule(lr){14-16}
 & \textbf{T-BN} & \textbf{S-BN} & \textbf{EN}
 & \textbf{T-BN} & \textbf{S-BN} & \textbf{EN}
 & \textbf{T-BN} & \textbf{S-BN} & \textbf{EN}
 & \textbf{T-BN} & \textbf{S-BN} & \textbf{EN}
 & \textbf{T-BN} & \textbf{S-BN} & \textbf{EN} \\
\midrule
Cardiology        & 50.00 & 58.33 & 66.67 & 75.00 & \cellcolor{cyan!10}\textbf{83.33} & 75.00 & 58.33 & 66.67 & 75.00 & 41.67 & 66.67 & 75.00 & 4.12 & 4.21 & 22.73 \\
Dentistry/Oral    & 56.25 & 62.50 & 68.75 & 56.25 & \cellcolor{cyan!10}\textbf{75.00} & 56.25 & 31.25 & 37.50 & 56.25 & 25.00 & 56.25 & 50.00 & 3.44 & 3.66 & 20.82 \\
Dermatology       & 29.81 & 39.42 & 40.38 & 36.54 & \cellcolor{cyan!10}\textbf{53.85} & 39.42 & 16.35 & 26.92 & 38.46 & 25.00 & 51.92 & \cellcolor{cyan!10}\textbf{53.85} & 1.25 & 1.65 & 14.72 \\
Endocrinology     & 75.00 & \cellcolor{cyan!10}\textbf{87.50} & 81.25 & 50.00 & 62.50 & 50.00 & 75.00 & 75.00 & \cellcolor{cyan!10}\textbf{87.50} & 50.00 & 68.75 & 75.00 & 3.19 & 3.47 & 19.57 \\
Gastroenterol.    & 41.67 & \cellcolor{cyan!10}\textbf{57.14} & \cellcolor{cyan!10}\textbf{57.14} & 35.71 & 50.00 & 42.86 & 28.57 & 35.71 & 50.00 & 28.57 & 50.00 & 50.00 & 2.41 & 2.69 & 18.75 \\
Gynecol./OB       & 48.00 & 60.00 & \cellcolor{cyan!10}\textbf{64.00} & 52.00 & \cellcolor{cyan!10}\textbf{64.00} & 56.00 & 24.00 & 28.00 & 44.00 & 20.00 & 40.00 & 44.00 & 1.88 & 2.44 & 15.12 \\
Neurology         & 50.00 & 56.25 & 62.50 & 62.50 & \cellcolor{cyan!10}\textbf{75.00} & 68.75 & 37.50 & 43.75 & 62.50 & 37.50 & 62.50 & 56.25 & 4.02 & 4.26 & 22.52 \\
Oncology          & \cellcolor{cyan!10}\textbf{100.00} & \cellcolor{cyan!10}\textbf{100.00} & \cellcolor{cyan!10}\textbf{100.00} & 75.00 & 83.33 & 75.00 & 0.00 & 8.33 & 33.33 & 50.00 & 66.67 & 75.00 & 0.94 & 1.23 & 16.21 \\
Ophthalmology     & 75.00 & \cellcolor{cyan!10}\textbf{83.33} & 75.00 & 75.00 & \cellcolor{cyan!10}\textbf{83.33} & 75.00 & 75.00 & 75.00 & \cellcolor{cyan!10}\textbf{83.33} & 0.00 & 41.67 & 41.67 & 2.11 & 2.29 & 18.96 \\
Orthopedics       & 22.22 & 33.33 & 38.89 & 50.00 & \cellcolor{cyan!10}\textbf{66.67} & 55.56 & 22.22 & 33.33 & 44.44 & 33.33 & 55.56 & 61.11 & 4.56 & 5.22 & 24.35 \\
Pediatrics        & 39.58 & 50.00 & 52.08 & 41.67 & \cellcolor{cyan!10}\textbf{58.33} & 47.92 & 8.33 & 18.75 & 37.50 & 27.08 & 43.75 & 47.92 & 1.04 & 1.55 & 13.10 \\
Psychiatry        & 25.00 & 41.67 & 33.33 & 75.00 & \cellcolor{cyan!10}\textbf{83.33} & \cellcolor{cyan!10}\textbf{83.33} & 0.00 & 8.33 & 33.33 & 0.00 & 25.00 & 25.00 & 3.81 & 3.88 & 24.45 \\
Pulmonology       & 33.33 & 50.00 & 50.00 & 58.33 & \cellcolor{cyan!10}\textbf{75.00} & 66.67 & 33.33 & 41.67 & 58.33 & 25.00 & 50.00 & 41.67 & 1.44 & 1.57 & 16.71 \\
Urology           & 58.33 & 66.67 & 66.67 & 75.00 & \cellcolor{cyan!10}\textbf{83.33} & \cellcolor{cyan!10}\textbf{83.33} & 16.67 & 25.00 & 50.00 & 58.33 & 75.00 & \cellcolor{cyan!10}\textbf{83.33} & 0.88 & 1.12 & 13.29 \\
Other             & 28.85 & 41.35 & 43.27 & 45.19 & 65.38 & 50.00 & 16.35 & 50.96 & 58.65 & 23.08 & \cellcolor{cyan!10}\textbf{83.65} & 82.69 & 3.14 & 3.51 & 19.95 \\
\midrule
\textbf{Average}  & 48.87 & 59.17 & 60.00 & 57.55 & \cellcolor{cyan!10}\textbf{70.83} & 61.67 & 29.53 & 38.33 & 54.17 & 29.64 & 55.83 & 57.50 & 2.55 & 2.85 & 18.75 \\
\bottomrule
\end{tabular}
\caption{Fine-grained performance breakdown (Avg.\ LAVE Score) across medical specialties for open-source and specialized models. The evaluation highlights the performance of Transliterated Bangla (\textbf{T-BN}), Standard Bangla (\textbf{S-BN}) and English (\textbf{EN}) queries.}
\label{tab:lave_opensource_models}
\end{table*}

\begin{figure*}[t]
    \centering
    \includegraphics[width=\textwidth]{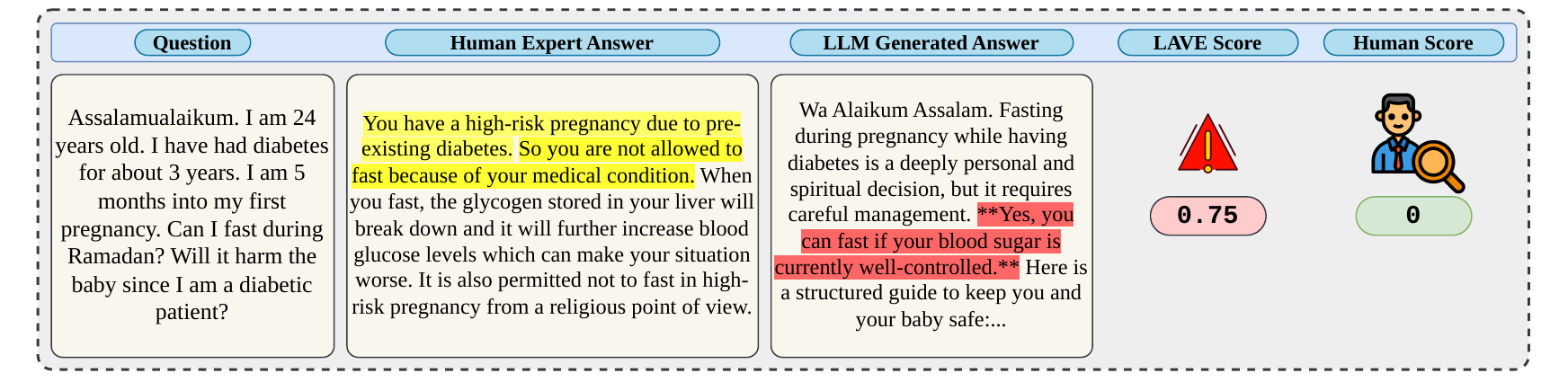} 
    \caption{Error analysis of medical question answering from the proposed dataset, highlighting score discrepancies between LLM judge and human expert.}
    \label{fig:error}
\end{figure*}

We present the results of BERTScore (BScore), and LLM Assisted VQA Evaluation (LAVE) and Human annotations in Table~\ref{tab:medical_vqa_results}. Table~\ref{tab:lave_proprietary_models},~\ref{tab:lave_opensource_models} reports the LAVE scores for different categories.

\noindent \textbf{Overall Performance.} Table~\ref{tab:medical_vqa_results} reports results under all three
metrics. Claude-Sonnet-4.6 and Gemini-3-Flash are the
strongest models overall, with Claude leading on Transliterated Bangla (57.53\% human) and Gemini on Standard
Bangla and English (70.34\% and 73.77\%). LAVE agrees with human evaluation on the broad ordering of models,
placing the closed-source systems at the top and LLaVA-Med at the bottom, but the two metrics disagree on which
model leads Standard Bangla and English. Among open-source systems,
Gemma-4-E4B-it is the strongest, achieving the highest BScore on Standard Bangla (46.10\%) among all models
evaluated and the highest human scores on Transliterated and Standard Bangla within its group. The two
medical-specialised models do not outperform general-purpose models of comparable scale: MedGemma-4B trails
Gemma-4-E4B-it on Transliterated Bangla under both human (20.60\% vs.\ 42.30\%) and LAVE (29.64\% vs.\ 57.55\%)
evaluation, and LLaVA-Med scores below 6\% human on all three query sets.

\noindent \textbf{Impact of Query Language.} Performance on Transliterated Bangla is lower than on Standard
Bangla and English for every model under both human and LAVE evaluation, with the sole exception of LLaVA-Med,
whose scores are near zero throughout. 
These three sets are each drawn independently from real patient queries, so differences in script co-occur with differences in query content, complexity and clinical framing. Table~\ref{tab:medical_vqa_results} and Tables~\ref{tab:lave_proprietary_models} and~\ref{tab:lave_opensource_models} therefore describe performance on naturally occurring query distributions and do not by themselves isolate the effect of script. We isolate script in the reformulation experiment below, which holds clinical content fixed by rewriting the same transliterated queries into Standard Bangla and English; conclusions about script sensitivity rest on that controlled comparison.

Tables~\ref{tab:lave_proprietary_models} and~\ref{tab:lave_opensource_models} break performance down by specialty. The difference between Transliterated and Standard Bangla is largest in the best-represented categories: on Dermatology, Claude-Sonnet-4.6 scores 53.85 on Transliterated Bangla against 64.42 on Standard Bangla, and Qwen3-VL-8B 29.81 against 39.42. Gynecology/OB and Pediatrics show the same pattern across most models. Endocrinology and Ophthalmology remain high and stable across all three query sets for the stronger models, while Psychiatry and Oncology vary widely, with several models scoring very low on Transliterated Bangla.

\noindent \textbf{Contribution of the image.} Table~\ref{tab:unimodal-multimodal} compares the image-bearing and text-only partitions, but these contain different queries and cannot show what the image adds. We therefore evaluate every image-bearing query twice, with its image and with the identical text alone (Table~\ref{tab:paired-image}). The image helps the closed-source models substantially in all three language conditions, adding on average 26.5 points on Transliterated Bangla, 24.1 on Standard Bangla and 29.0 on English. Gains are smaller and less consistent for the open and medical models: Qwen3-VL-8B, Gemma-4-E4B-it and MedGemma-4B improve in every condition, Llama-3.2-11B-Vision gains 25.0 points on Transliterated Bangla but loses slightly on Standard Bangla and English, and LLaVA-Med-v1.5 changes little.

\noindent \textbf{Effect of Query Reformulation.} To test whether transliteration itself is what depresses Transliterated Bangla performance, we rewrite the original transliterated queries into Standard Bangla and English and re-evaluate every model; the reformulation procedure and its validation are detailed in Appendix~\ref{app:reformulation} and full per-specialty results in Appendix Tables~\ref{tab:lave_proprietary_improvement},~\ref{tab:lave_opensource_improvement}
and~\ref{tab:lave_medspecialized_improvement}. The effect divides the models sharply along one line: how much they were losing to transliteration in the first place. The open-source and medical models, which suffered most, recover the most. Llama-3.2-11B climbs from 29.53 to 54.32 under English reformulation, MedGemma-4B from 29.64 to 51.60 and Qwen3-VL-8B from 48.87 to 61.51, each gaining 12 to 25 LAVE points from a change of script alone. The closed-source models, by contrast, scarcely move: GPT-5-Mini gains 9.05 points in English, while Claude-Sonnet-4.6 and Gemini-3-Flash actually edge down. They had already read the transliterated queries at close to full strength, so there was little left for reformulation to recover. Standard Bangla reformulation helps almost every model but by less than English, and for MedGemma-4B nearly as much (+21.85 vs.\ +21.96). LLaVA-Med is unmoved throughout, gaining under 5 points even in English. Physician ratings on a 200-query subset confirm this pattern (Appendix Table~\ref{tab:reform_human}): the score changes from reformulation under human and LAVE evaluation agree closely (Pearson $r = 0.90$ across the 16 model--language changes), with the largest human gains again for Llama-3.2-11B-V ($+24.65$ in English), MedGemma-4B ($+19.20$) and Qwen3-VL-8B ($+14.80$), and little change for Claude-Sonnet-4.6 and Gemini-3-Flash. The one notable difference is GPT-5-Mini, whose English gain is much larger under human evaluation ($+22.05$) than under LAVE ($+8.35$).

\noindent \textbf{LLM Judges Overrate Responses.} A notable discrepancy is observed between LAVE scores and human evaluation judgments throughout Table~\ref{tab:medical_vqa_results}. LAVE consistently inflates scores relative to human assessors across all eight models on Standard Bangla, with the largest gaps for GPT-5-Mini (80.17\% LAVE vs.\ 57.60\% human, a 22-point difference) and Claude-Sonnet-4.6 (82.50\% vs.\ 65.70\%). Figure~\ref{fig:error} presents a detailed error analysis of this discrepancy. The divergence is consistent with the self-enhancement bias documented in LLM-based evaluation~\cite{zheng2023judging, wang-etal-2024-large-language-models-fair}: judge models tend to reward fluent, verbose, and confidently stated outputs stylistic properties correlated with model-generated text even when the clinical content is incomplete or subtly incorrect. Human annotators, in contrast, are sensitive to factual omissions, hallucinated findings, and clinically inappropriate recommendations that a language-model judge may overlook. This bias is especially consequential in the medical domain, where a fluent but erroneous answer is potentially more harmful than an incomplete one.
To quantify how far LAVE can be trusted, we compute rank correlations between LAVE and the mean physician score over the 600-response human-evaluation subset (Table~\ref{tab:lave_human_corr}). Correlation is moderate and significant in every language condition (overall Spearman $\rho = 0.53$, Kendall $\tau = 0.41$, $p < 0.001$). LAVE is therefore useful for relative comparison of models at scale, but it is not a substitute for clinical judgment.

\begin{table}[t]
\centering
\small
\setlength{\tabcolsep}{3pt}
\begin{tabular}{l ccc}
\toprule
\textbf{Language} & \textbf{Spearman $\rho$} & \textbf{Kendall $\tau$} & \textbf{$p$} \\
\midrule
Transliterated Bangla & 0.48 & 0.38 & $<0.001$ \\
Standard Bangla       & 0.55 & 0.43 & $<0.001$ \\
English               & 0.56 & 0.44 & $<0.001$ \\
\midrule
\textbf{Overall}      & \textbf{0.53} & \textbf{0.41} & $<0.001$ \\
\bottomrule
\end{tabular}
\caption{Rank correlation between LAVE and physician scores ($n = 600$; 200 per language).}
\label{tab:lave_human_corr}
\end{table}

\section{Conclusion}
We introduce the first open-ended, multimodal medical question-answering benchmark for the Bangla language ecosystem, built from authentic transliterated, Bangla, and English patient queries sourced from online healthcare communities in Bangladesh. Evaluating leading proprietary, open-source, and medical-specialised models exposes a sharp divide on informal transliterated input: closed-source models such as Claude-Sonnet-4.6 answer these queries with only a modest drop, while open-source and domain-specialised models degrade sharply. Our counterfactual reformulation experiment locates the cause: rewriting transliterated queries into Standard Bangla or English recovers much of the loss for the weaker models (Llama-3.2-11B and MedGemma-4B each gain over twenty LAVE points in English), showing their deficit is largely surface-level script rather than clinical reasoning, whereas the frontier models gain little because script was never their bottleneck. Finally, LAVE consistently scores responses above human clinicians, revealing a self-enhancement bias that rewards fluent but sometimes clinically incomplete answers. Together these findings set a more realistic standard for medical AI in low-resource settings and underscore that human verification remains necessary before deployment.

\section*{Limitations}

Despite the contributions of this work, several limitations remain. Although overall model performance improves under standard language and English reformulations, accuracy on fine-grained and visually nuanced categories such as Psychiatry and Oncology remains highly volatile under noisy, transliterated inputs. Two constraints apply specifically to the reformulation experiment. First, the automatic reformulations were validated by human annotators on a 500-example subset rather than exhaustively. Second, physician ratings of the reformulated queries cover a 200-query subset rather than the full set. Reference answers were written and validated by MBBS-qualified general practitioners rather than specialists. This suits the consumer-level queries in the dataset, but the references should not be read as specialist consensus, and specialist review of high-risk categories such as Oncology, Psychiatry and Cardiology is left to future work. Several specialty categories are also small, so per-specialty scores carry higher variance than the macro-averaged overall scores. Finally, while the dataset spans multiple specialist medical domains, automated LLM-as-a-judge frameworks continue to demonstrate a persistent self-enhancement bias, overrating fluent but clinically incomplete answers and underscoring the ongoing necessity of expensive human clinical verification.

\section*{Ethical Statement}
This work involves the creation and use of an open-ended, multimodal medical question-answering dataset in Bangla. All clinical queries and associated images were sourced from publicly available online medical communities in Bangladesh. Patient data privacy has been rigorously maintained throughout; a meticulous filtering process was executed to scrub all text-based personally identifiable information and any medical imagery revealing patient identities or distinct facial features was blurred or permanently removed. Annotations, gold-standard reference generation and dataset validation tiers were performed entirely by qualified, MBBS-certified medical practitioners. We acknowledge that models evaluated on this dataset are intended for research and benchmarking purposes only and should not be deployed for autonomous clinical decision making. Additionally, LLMs were utilized during the drafting process to assist with grammatical refinement and structural clarity; however, all technical content and final interpretations were produced and verified by the authors.

\bibliography{custom}

\appendix

\section{Dataset Annotation and Compensation Details}
\label{app:compensation}

Annotators were given a two-month deadline to complete the answer-generation
task for the final set of medical queries. All reference answers were written
in standard English regardless of the language or script of the original
query, so that model responses from the transliterated Bangla, standard
Bangla, and English query sets could be evaluated against a common reference
answer.

\paragraph{Compensation.}
Annotators received 20~BDT per reference answer and validators received
5~BDT per rating. Writing one reference answer took approximately 2--3
minutes on average, corresponding to an effective rate of 400--600~BDT per
hour. A single validation rating took approximately 1 minute on average,
corresponding to an effective rate of approximately 300~BDT per hour. Physicians who rated model responses for the human evaluation, including the additional physician who rated the reformulation subset, were compensated at the same per-rating rate as the validators.

\paragraph{Benchmark.}
Since all annotators and validators are MBBS-qualified physicians, we
benchmark their compensation against Grade~9 of the Bangladesh National Pay
Scale 2026, the entry grade for officers recruited through the Bangladesh
Civil Service (BCS), including the health cadre through which physicians
enter government service.\footnote{Service (Pay and Allowances) Order, 2026,
S.R.O. No.~347-Law/2026, Bangladesh Gazette (Extraordinary), Ministry of
Finance, 17 September 2026:
\url{https://www.dpp.gov.bd/bgpress/index.php/document/get_extraordinary/62975}}
Once the scale is fully implemented, an entry-level Grade~9 officer receives
a basic salary of 44{,}000~BDT per month and a total monthly pay of
approximately 60{,}000~BDT including allowances. Assuming a standard 40-hour
working week ($\approx$173 hours per month), this total corresponds to
approximately 346~BDT per hour.
The answer-writing rate exceeds this benchmark by 16--73\%, while the
validation rate is slightly below it (approximately 87\% of the benchmark).
Because the same physicians performed both tasks, their average effective
rate across all annotation and validation work was approximately
360--450~BDT per hour, i.e., 4--30\% above the hourly equivalent of the
full entry-level pay of a government-employed physician.

\paragraph{Information given to annotators.}
Before starting, all annotators, validators and evaluators were informed of the purpose of the study and how their answers and ratings would be used and released. They were told that the queries were real patient posts from public online communities and could describe distressing medical situations, and that their work was for research only and would not be used to give advice to the original posters. Participation was voluntary; they could skip any query they were uncomfortable answering and could withdraw at any time without penalty, with payment for all work completed up to that point. No personal information about the annotators is included in the dataset, and all agreed to these terms before beginning.

\section{Annotation and Validation Guidelines}
\label{appendix:guidelines}

This section details the guidelines provided to annotators and validators during 
the answer generation and validation stages. All annotators and validators were 
MBBS-qualified general practitioners.

\subsection{Annotator Guidelines (Answer Generation)}

Annotators were tasked with generating reference (gold) answers of each query in English. They were instructed to ensure that each answer satisfied the following criteria:

\begin{itemize}
    \item \textbf{Accuracy.} Answers must be factually correct and consistent with 
    current medical consensus and standard clinical practice.
    \item \textbf{Completeness.} Answers should include all clinically necessary 
    information to address the query, while avoiding detail unrelated to the question.
    \item \textbf{Relevance.} Answers must directly address the question asked.
    \item \textbf{Explanation.} Where appropriate, answers should include a brief 
    rationale rather than an unexplained conclusion, as explanations improve the 
    clinical usefulness and trustworthiness of an answer.
\end{itemize}

Annotators were further instructed to (i) answer only within their assigned medical 
domain; (ii) explicitly advise in-person care when a condition was potentially 
serious, time-sensitive, or required physical examination; and (iii) avoid fabricating 
assumptions when a query lacked sufficient information, instead stating what 
additional information would be required.

\subsection{Validator Guidelines (Answer Validation)}

For answering the queries assigned to them, validators followed the same guidelines 
as the annotators. For the agreement rating, they then rated the answer on a 
two-point scale:

\begin{itemize}
    \item \textbf{0 (Disagree):} the answer is inaccurate, clinically 
    incomplete, or does not address the question.
    \item \textbf{1 (Agree):} the answer is medically correct, adequately complete and the validator would be comfortable providing it.
\end{itemize}

Validators were instructed to base their ratings on clinical correctness and adequacy 
rather than writing style, so that the rating of 0 was reserved for answers that were genuinely incorrect or non-relevant.

\begin{figure*}[t]
  \centering
  \includegraphics[width=\textwidth]{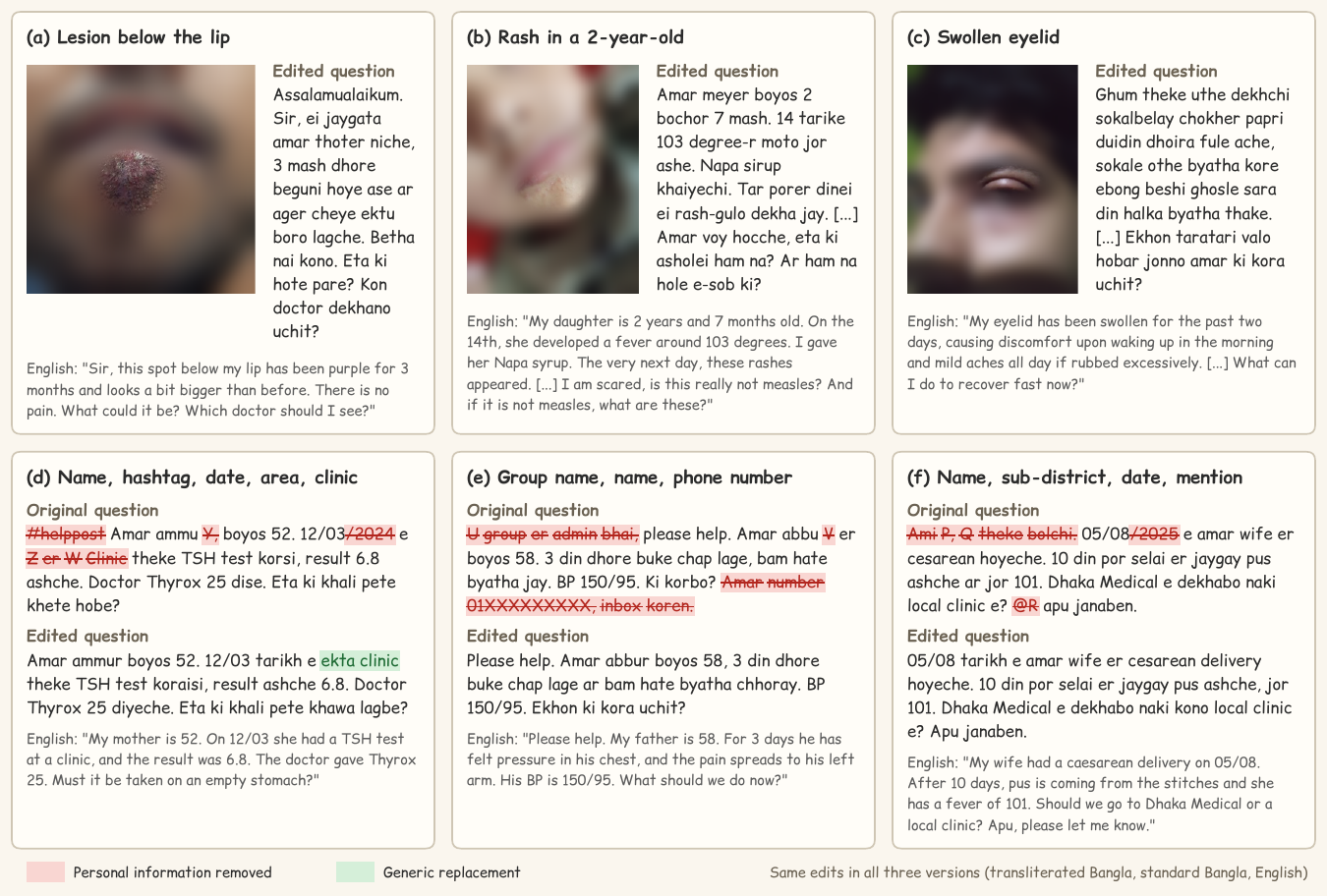}
  \caption{Examples of anonymisation in BanTMedQA. (a)--(c) Images are blurred except for the affected area, and the paired question is lightly paraphrased; [...] marks text omitted here for space. (d)--(f) Personal details are removed or replaced with generic terms, and the query is lightly paraphrased. In the original questions, capital letters (P, Q, R, U, V, W, X, Y, Z) and 01XXXXXXXXX are placeholders for names, places, a group name and a phone number, so that no personal information appears in this figure. All edits are applied in the same way to the transliterated Bangla, standard Bangla and English versions.}
  \label{fig:anonymisation}
\end{figure*}

\section{Query Reformulation Procedure}
\label{app:reformulation}

For the counterfactual reformulation experiment, we re-expressed each original
Transliterated Bangla query in two target forms, Standard Bangla and English, while holding the clinical content of the query fixed. Reformulation was performed automatically with \texttt{Gemini-3-Flash}, prompted to convert the query into the target language without adding, removing, or altering any clinical detail. The paired image, where present, and the expert reference answer were left unchanged, so that a model's score on a reformulated query is directly comparable to its score on the original.

\paragraph{Validation.} To confirm that reformulation preserved meaning rather than silently altering the queries, we drew a random subset of 500 reformulated queries and had two annotators
independently judge each one against its original transliterated form, marking whether the reformulation was faithful, i.e.\ it preserved the clinical intent and all salient details of the source query without introducing new information. Both annotators judged 98.0\% of the subset identically: 490 reformulations were marked faithful by both, none were marked unfaithful by both, and 10 were disagreements. As in dataset validation, the marginal distribution is extremely skewed (99.0\% of judgments were ``faithful''), so Cohen's $\kappa$ is uninformative here ($\kappa = -0.010$, 95\% CI $[-0.016, -0.004]$). We report Gwet's AC$_1 = 0.980$ (95\% CI $[0.967, 0.992]$) and PABAK $= 0.960$; full counts are given in Table~\ref{tab:agree-reform}. The substantive conclusion rests on the observed rate: 98.0\% of reformulations were judged faithful by both annotators, and none was judged unfaithful by both.

\section{Human Evaluation Protocol}\label{app:human_eval}

This appendix details the protocol used to obtain the human evaluation scores reported in Table~\ref{tab:medical_vqa_results}.

\paragraph{Evaluators.}
Human evaluation was carried out by 2 MBBS-qualified general practitioners, each of whom had completed an MBBS degree from a medical college in Bangladesh followed by a one-year internship. The evaluation subset contains 600 queries (200 per language condition), and both evaluators independently rated every model response in it. The physician ratings for the script reformulation subset (Table~\ref{tab:reform_human}) were provided by an additional MBBS-qualified general practitioner with the same training, using the same rubric.

\paragraph{Agreement.}
Because the scale is ordinal, we measure inter-rater reliability with quadratic-weighted Cohen's $\kappa$. The evaluators gave the same rating on 62.5\% of responses and ratings within one adjacent level on 93.8\%, with $\kappa_w = 0.76$. Disagreements were not adjudicated to consensus: the final score of each response is the mean of the two ratings, which is then macro-averaged over medical categories, matching the aggregation used for LAVE. The overall 95\% confidence interval for human scores is approximately $\pm 0.08$.

\paragraph{Rating scale.}
Evaluators rated each model response against the corresponding expert reference answer on the following five-point scale, identical to the rubric supplied to the LAVE judge (Appendix~\ref{appendix: lave_score}):

\begin{itemize}
    \item \textbf{1.0} -- Completely correct: same meaning as reference.
    \item \textbf{0.75} -- Mostly correct: minor differences only.
    \item \textbf{0.5} -- Partially correct: some correct elements, significant errors.
    \item \textbf{0.25} -- Mostly incorrect: some relevant content but largely wrong.
    \item \textbf{0.0} -- Completely incorrect: wrong or irrelevant.
\end{itemize}

Evaluators were instructed to base their judgments on clinical correctness, relevance, and completeness relative to the reference answer, and explicitly not to penalise responses for stylistic or lexical divergence where the clinical content remained accurate and complete. Responses containing harmful medication advice, a missed red flag requiring urgent referral, or hallucinated clinical information were assigned to the lowest score tiers regardless of how fluent or complete the rest of the response was.

\section{Medical Category Distribution}
\label{app:category_sizes}

Table~\ref{tab:category_sizes} lists the number of queries in each of the 15 medical categories of \name{}. The distribution reflects how often each specialty appears in the source communities: Dermatology, Gynecology/Obstetrics and the ``Other'' category together account for over 60\% of the queries, while Neurology, Oncology, Ophthalmology and Psychiatry each contain fewer than 20. Overall scores are macro-averaged across categories so that the largest categories do not dominate them, but per-category scores for the smallest categories in Tables~\ref{tab:lave_proprietary_models} and~\ref{tab:lave_opensource_models} rest on few queries and should be interpreted with caution.

\begin{table}[h]
\centering
\small
\begin{tabular}{l r}
\toprule
\textbf{Category} & \textbf{Queries} \\
\midrule
Cardiology      & 45 \\
Dentistry/Oral  & 153 \\
Dermatology     & 451 \\
Endocrinology   & 45 \\
Gastroenterol.  & 81 \\
Gynecol./OB     & 343 \\
Neurology       & 18 \\
Oncology        & 18 \\
Ophthalmology   & 10 \\
Orthopedics     & 117 \\
Pediatrics      & 144 \\
Psychiatry      & 18 \\
Pulmonology     & 27 \\
Urology         & 36 \\
Other           & 298 \\
\midrule
\textbf{Total}  & \textbf{1{,}804} \\
\bottomrule
\end{tabular}
\caption{Number of queries per medical category in \name{}.}
\label{tab:category_sizes}
\end{table}

\section{Inter-Annotator Agreement Statistics}
\label{app:agreement}

Both raters assigned the ``agree'' label to 95.0\% of pairs. Under marginals
this skewed, the chance-agreement term in Cohen's $\kappa$ ($P_e = 0.905$)
slightly exceeds the observed agreement of 0.904, yielding
$\kappa = -0.011$ (95\% CI $[-0.050, 0.029]$) despite raters concurring on
nine pairs in ten. This is the first Feinstein--Cicchetti paradox, in which
$\kappa$ collapses under extreme prevalence skew and ceases to reflect
observed concordance \citep{feinstein1990high,byrt1993bias}. We therefore
report Gwet's AC$_1 = 0.894$ (95\% CI $[0.878, 0.910]$) and PABAK $= 0.808$
alongside raw observed agreement. We do not claim high chance-corrected
agreement: given these marginals, rater concordance cannot be distinguished
from chance by $\kappa$. The retention rule is conservative independently of
the agreement statistic, since a pair enters the final dataset only under
unanimous endorsement, so residual disagreement removes items rather than
admitting them. Table~\ref{tab:agree-validation} gives full contingency
counts for dataset validation and Table~\ref{tab:agree-reform} for
reformulation faithfulness.

\begin{table}[t]
\centering
\small
\begin{tabular}{lrrr}
\toprule
& \textbf{Rater B $=1$} & \textbf{Rater B $=0$} & \textbf{Total} \\
\midrule
\textbf{Rater A $=1$} & 1{,}804 & 96  & 1{,}900 \\
\textbf{Rater A $=0$} & 96      & 4   & 100 \\
\midrule
\textbf{Total}        & 1{,}900 & 100 & 2{,}000 \\
\bottomrule
\end{tabular}

\vspace{0.6em}

\begin{tabular}{lr}
\toprule
\textbf{Statistic} & \textbf{Value} \\
\midrule
Observed agreement $P_o$      & 0.904 [0.891, 0.917] \\
Expected agreement $P_e$ ($\kappa$) & 0.905 \\
Marginal $p(1)$: Rater A / B  & 0.950 / 0.950 \\
Cohen's $\kappa$              & $-0.011$ $[-0.050, 0.029]$ \\
Gwet's AC$_1$                 & 0.894 [0.878, 0.910] \\
PABAK                         & 0.808 [0.782, 0.834] \\
Prevalence index              & 0.900 \\
Bias index                    & 0.000 \\
\bottomrule
\end{tabular}
\caption{Contingency table and agreement statistics for dataset validation
($n = 2{,}000$). Cohen's $\kappa$ is depressed by extreme prevalence skew;
Gwet's AC$_1$ and raw observed agreement are reported alongside it.}
\label{tab:agree-validation}
\end{table}

\begin{table}[t]
\centering
\small
\begin{tabular}{lrrr}
\toprule
& \textbf{B: Faithful} & \textbf{B: Unfaith.} & \textbf{Total} \\
\midrule
\textbf{A: Faithful}   & 490 & 5 & 495 \\
\textbf{A: Unfaithful} & 5   & 0 & 5 \\
\midrule
\textbf{Total}         & 495 & 5 & 500 \\
\bottomrule
\end{tabular}

\vspace{0.6em}

\begin{tabular}{lr}
\toprule
\textbf{Statistic} & \textbf{Value} \\
\midrule
Observed agreement $P_o$      & 0.980 [0.968, 0.992] \\
Expected agreement $P_e$ ($\kappa$) & 0.980 \\
Marginal $p(\text{faithful})$: A / B & 0.990 / 0.990 \\
Cohen's $\kappa$              & $-0.010$ $[-0.016, -0.004]$ \\
Gwet's AC$_1$                 & 0.980 [0.967, 0.992] \\
PABAK                         & 0.960 [0.936, 0.985] \\
Prevalence index              & 0.980 \\
Bias index                    & 0.000 \\
\bottomrule
\end{tabular}
\caption{Contingency table and agreement statistics for reformulation
faithfulness ($n = 500$). Observed agreement is 98.0\%; $\kappa$ is
uninformative under near-unanimous marginals.}
\label{tab:agree-reform}
\end{table}

\section{Data Privacy and Release}
\label{app:release}

We release BanTMedQA to support reproducibility and further research on
medical question answering in low-resource, multilingual settings. The dataset is distributed under gated access with a Data Use Agreement (DUA) rather than as an open download.

\paragraph{Hosting and format.}
The dataset is hosted as a gated dataset on the Hugging Face Hub
(\url{https://huggingface.co/datasets/FaiyazAbdullah114708/BanTMedQA}) and, once access is granted, can be loaded through
the \texttt{datasets} library. Each example is provided as a structured
record containing the query in its available script variants
(transliterated Bangla, standard Bangla, and English), the query type
(direct or story-based), the medical-category label, the associated clinical
image where present, the physician-written English reference answer.
All embedded image metadata (e.g., EXIF) has been removed.

\paragraph{Access procedure.}
Access requests are reviewed manually by the authors. Applicants must
provide their name, institutional affiliation, an institutional email
address, and a brief description of their intended research use, and must
accept the DUA below. Requests from anonymous accounts or for purposes
inconsistent with the DUA are declined. We keep a record of approved users
and revoke access if the DUA is violated.

\paragraph{Data Use Agreement.}
By requesting access, users agree to:
\begin{itemize}
  \item use the dataset only for non-commercial research and education;
  \item make no attempt to re-identify, locate, or contact any individual who
        authored a query or appears in an image, including by searching for
        query text or images on social media platforms or search engines;
  \item not redistribute the dataset, or any derived version containing the
        original query text or images, and direct other researchers to the
        official access page instead;
  \item not use the dataset, or models trained or tuned on it, for autonomous
        clinical decision-making or to provide medical advice to patients;
  \item report any personally identifying content discovered in the dataset
        to the authors and not use or disclose it further;
  \item delete specific items on notice from the authors, and delete all
        copies of the dataset if access is revoked.
\end{itemize}
Short excerpts may be quoted in publications for illustration, provided they
contain no identifying information.

\paragraph{License.}
Subject to the DUA, our own contributions are licensed under CC BY-NC-SA 4.0.
These include the physician-written reference answers, the category and
query-type labels, the query reformulations. The query text and images were originally written or
shared by members of the public. We claim no ownership of them and provide
them solely for non-commercial research under the terms of the DUA.

\paragraph{Privacy and anonymization.} Queries were filtered and anonymized as described in \S\ref{sec:dataset}. Each paraphrased query was then checked against its original to confirm that its clinical content was unchanged. As a check against search-based tracing, we searched a random sample of 50 paraphrased queries as exact strings on Facebook and Google; none returned the source post. Images were additionally checked with reverse image search and none returned the source post. The release contains no source group names, usernames, profile links, post URLs, post identifiers, or post timestamps.

\paragraph{Takedown requests.}
Anyone who believes their content appears in the dataset may request its
removal by contacting \texttt{[msayeedi212049@bscse.uiu.ac.bd]}. Confirmed items are removed
from the hosted dataset, and approved users are notified to delete them
under the terms of the DUA.

\paragraph{Intended use.}
The dataset is intended solely for benchmarking and research on medical
question answering. It must not be used for autonomous clinical
decision-making or to provide medical advice. Models evaluated on it are not
validated for clinical deployment.

\section{Used Prompts in the Paper} \label{all_prompts}
\begin{center}
\begin{tcolorbox}[
    colback=gray!5!white,
    colframe=black,
    title={Structured Prompt for Model Inference},
    width=0.49\textwidth
]
\small

\textbf{System Prompt:}\\[2pt]
You are a knowledgeable medical assistant. Answer the patient's question concisely and accurately. Use 2--5 short sentences unless the question requires more detail. If an image is provided, analyse it carefully as part of your response. Do not refuse to answer medical questions; provide helpful, evidence-based information. Answer the user's question strictly in English. Do not answer in Bangla, Bengali script, or transliteration.\\[6pt]

\textbf{User Prompt Template:}\\[2pt]
\textbf{Question:}\\
\texttt{\{question\}}\\[6pt]

\textbf{Output Requirements:}
\begin{itemize}
    \item Answer only in English.
    \item Do not use Bangla, Bengali script, or transliteration.
    \item Do not translate the question; answer it directly.
    \item Keep the answer concise.
\end{itemize}

\end{tcolorbox}
\end{center}

\noindent \textbf{LAVE Evaluation Prompt}
\label{appendix: lave_score}
\begin{center}
\begin{tcolorbox}[
    colback=gray!5!white,
    colframe=black,
    title={Structured Prompt for LAVE Evaluation},
    width=0.48\textwidth
]
\small

You are a medical answer evaluation expert. You will receive a numbered list of
\texttt{(Reference Answer, Generated Answer)} pairs. For each pair, score the generated
answer on a scale from 0 to 1:\\[4pt]

\begin{itemize}
    \item \textbf{1.0} -- Completely correct: same meaning as reference.
    \item \textbf{0.75} -- Mostly correct: minor differences only.
    \item \textbf{0.5} -- Partially correct: some correct elements, significant errors.
    \item \textbf{0.25} -- Mostly incorrect: some relevant content but largely wrong.
    \item \textbf{0.0} -- Completely incorrect: wrong or irrelevant.
\end{itemize}

\vspace{4pt}
Respond with \textbf{only} a valid JSON array of numbers: one float per pair, in the same order.\\[4pt]

\textbf{Example for three pairs:}
\begin{verbatim}
[1.0, 0.5, 0.25]
\end{verbatim}

Do \textbf{not} include any text outside the JSON array.

\end{tcolorbox}
\end{center}

\begin{table*}[ht]
\centering
\setlength{\tabcolsep}{4pt}
\small
\begin{tabular}{l ccc ccc ccc}
\toprule
\multirow{2}{*}{\textbf{Category}}
  & \multicolumn{3}{c}{\textbf{Claude-Sonnet-4.6}}
  & \multicolumn{3}{c}{\textbf{Gemini-3-Flash}}
  & \multicolumn{3}{c}{\textbf{GPT-5-Mini}} \\
\cmidrule(lr){2-4}\cmidrule(lr){5-7}\cmidrule(lr){8-10}
 & \textbf{T-BN} & \textbf{$\rightarrow$S-BN} & \textbf{$\rightarrow$EN}
 & \textbf{T-BN} & \textbf{$\rightarrow$S-BN} & \textbf{$\rightarrow$EN}
 & \textbf{T-BN} & \textbf{$\rightarrow$S-BN} & \textbf{$\rightarrow$EN} \\
\midrule
Cardiology
  & 83.33 & 83.33 & 58.33
  & 75.00 & \cellcolor{cyan!10}\textbf{91.67} & 66.67
  & 83.33 & \cellcolor{cyan!10}\textbf{91.67} & 83.33 \\

Dentistry/Oral
  & 68.75 & 75.00 & 75.00
  & 50.00 & 56.25 & 50.00
  & 75.00 & 62.50 & \cellcolor{cyan!10}\textbf{87.50} \\

Dermatology
  & 53.85 & 53.85 & \cellcolor{cyan!10}\textbf{57.69}
  & 50.00 & 52.88 & 52.88
  & 42.31 & 45.19 & 56.73 \\

Endocrinology
  & \cellcolor{cyan!10}\textbf{100.00} & 91.67 & 83.33
  & 91.67 & 83.33 & 75.00
  & 91.67 & 83.33 & 75.00 \\

Gastroenterol.
  & 60.71 & \cellcolor{cyan!10}\textbf{82.14} & \cellcolor{cyan!10}\textbf{82.14}
  & 53.57 & 60.71 & 67.86
  & 50.00 & 64.29 & 78.57 \\

Gynecol./OB
  & 77.00 & \cellcolor{cyan!10}\textbf{81.00} & 80.00
  & 65.00 & 73.00 & 73.00
  & 65.00 & 75.00 & \cellcolor{cyan!10}\textbf{81.00} \\

Neurology
  & 62.50 & 62.50 & 50.00
  & 37.50 & 37.50 & 25.00
  & \cellcolor{cyan!10}\textbf{75.00} & \cellcolor{cyan!10}\textbf{75.00} & \cellcolor{cyan!10}\textbf{75.00} \\

Oncology
  & 25.00 & 25.00 & 25.00
  & \cellcolor{cyan!10}\textbf{100.00} & \cellcolor{cyan!10}\textbf{100.00} & \cellcolor{cyan!10}\textbf{100.00}
  & \cellcolor{cyan!10}\textbf{100.00} & \cellcolor{cyan!10}\textbf{100.00} & \cellcolor{cyan!10}\textbf{100.00} \\

Ophthalmology
  & \cellcolor{cyan!10}\textbf{100.00} & \cellcolor{cyan!10}\textbf{100.00} & \cellcolor{cyan!10}\textbf{100.00}
  & 75.00 & 75.00 & 75.00
  & 75.00 & 75.00 & 75.00 \\

Orthopedics
  & 55.56 & \cellcolor{cyan!10}\textbf{66.67} & 63.89
  & 52.78 & 55.56 & 52.78
  & 50.00 & 58.33 & \cellcolor{cyan!10}\textbf{66.67} \\

Pediatrics
  & 58.33 & 54.17 & 56.25
  & 45.83 & 37.50 & 41.67
  & 37.50 & 54.17 & \cellcolor{cyan!10}\textbf{60.42} \\

Psychiatry
  & \cellcolor{cyan!10}\textbf{75.00} & \cellcolor{cyan!10}\textbf{75.00} & 25.00
  & \cellcolor{cyan!10}\textbf{75.00} & \cellcolor{cyan!10}\textbf{75.00} & 50.00
  & \cellcolor{cyan!10}\textbf{75.00} & \cellcolor{cyan!10}\textbf{75.00} & 25.00 \\

Pulmonology
  & 75.00 & \cellcolor{cyan!10}\textbf{83.33} & \cellcolor{cyan!10}\textbf{83.33}
  & 66.67 & 75.00 & \cellcolor{cyan!10}\textbf{83.33}
  & 41.67 & 50.00 & \cellcolor{cyan!10}\textbf{83.33} \\

Urology
  & 75.00 & \cellcolor{cyan!10}\textbf{83.33} & \cellcolor{cyan!10}\textbf{83.33}
  & 75.00 & 75.00 & 75.00
  & 50.00 & 66.67 & \cellcolor{cyan!10}\textbf{83.33} \\

Other
  & 74.04 & 76.92 & 75.96
  & 58.65 & 60.58 & 65.38
  & 62.50 & 68.27 & \cellcolor{cyan!10}\textbf{78.85} \\
\midrule
\textbf{Average}
  & 69.60 & 72.93$_{\textcolor{teal}{+3.32}}$ & 66.62$_{\textcolor{red}{-2.99}}$
  & 64.78 & 67.27$_{\textcolor{teal}{+2.49}}$ & 63.57$_{\textcolor{red}{-1.21}}$
  & 64.93 & 69.63$_{\textcolor{teal}{+4.70}}$ & \cellcolor{cyan!10}\textbf{73.98}$_{\textcolor{teal}{+9.05}}$ \\
\bottomrule
\end{tabular}
\caption{Counterfactual query reformulation results (Avg.\ LAVE Score) for closed-source models across medical specialties. \textbf{T-BN} denotes the original Transliterated Bangla queries; $\rightarrow$\textbf{S-BN} and $\rightarrow$\textbf{EN} denote reformulations into Standard Bangla and English respectively. Subscripts in the \textbf{Average} row show the absolute gain (\textcolor{teal}{green}) or loss (\textcolor{red}{red}) relative to T-BN. Cells shaded \colorbox{cyan!10}{cyan} mark the best score per specialty.}
\label{tab:lave_proprietary_improvement}
\end{table*}

\begin{table*}[ht]
\centering
\setlength{\tabcolsep}{4pt}
\small
\begin{tabular}{l ccc ccc ccc}
\toprule
\multirow{2}{*}{\textbf{Category}}
  & \multicolumn{3}{c}{\textbf{Qwen3-VL-8B}}
  & \multicolumn{3}{c}{\textbf{Gemma-4-E4B-it}}
  & \multicolumn{3}{c}{\textbf{Llama-3.2-11B-V}} \\
\cmidrule(lr){2-4}\cmidrule(lr){5-7}\cmidrule(lr){8-10}
 & \textbf{T-BN} & \textbf{$\rightarrow$S-BN} & \textbf{$\rightarrow$EN}
 & \textbf{T-BN} & \textbf{$\rightarrow$S-BN} & \textbf{$\rightarrow$EN}
 & \textbf{T-BN} & \textbf{$\rightarrow$S-BN} & \textbf{$\rightarrow$EN} \\
\midrule
Cardiology
  & 50.00 & 75.00 & 66.67
  & 75.00 & \cellcolor{cyan!10}\textbf{83.33} & 75.00
  & 58.33 & 58.33 & 75.00 \\

Dentistry/Oral
  & 56.25 & 56.25 & \cellcolor{cyan!10}\textbf{68.75}
  & 56.25 & 62.50 & 62.50
  & 31.25 & 50.00 & \cellcolor{cyan!10}\textbf{68.75} \\

Dermatology
  & 29.81 & 36.54 & 39.00
  & 36.54 & 37.50 & 42.31
  & 16.35 & 25.00 & 40.38 \\

Endocrinology
  & 75.00 & 75.00 & \cellcolor{cyan!10}\textbf{83.33}
  & 50.00 & 50.00 & 50.00
  & \cellcolor{cyan!10}\textbf{75.00} & 58.33 & 50.00 \\

Gastroenterol.
  & 41.67 & 42.86 & 57.14
  & 35.71 & 46.43 & 57.14
  & 28.57 & 39.29 & 57.14 \\

Gynecol./OB
  & 48.00 & 63.00 & \cellcolor{cyan!10}\textbf{65.00}
  & 52.00 & 59.00 & 59.00
  & 24.00 & 38.00 & 57.00 \\

Neurology
  & 50.00 & 62.50 & 62.50
  & 62.50 & 50.00 & 50.00
  & 37.50 & 62.50 & 50.00 \\

Oncology
  & \cellcolor{cyan!10}\textbf{100.00} & \cellcolor{cyan!10}\textbf{100.00} & \cellcolor{cyan!10}\textbf{100.00}
  & 75.00 & 75.00 & 75.00
  & 0.00 & 0.00 & 50.00 \\

Ophthalmology
  & 75.00 & 75.00 & 50.00
  & \cellcolor{cyan!10}\textbf{75.00} & \cellcolor{cyan!10}\textbf{75.00} & \cellcolor{cyan!10}\textbf{75.00}
  & \cellcolor{cyan!10}\textbf{75.00} & 50.00 & \cellcolor{cyan!10}\textbf{75.00} \\

Orthopedics
  & 22.22 & 30.56 & 33.33
  & 50.00 & \cellcolor{cyan!10}\textbf{52.78} & \cellcolor{cyan!10}\textbf{52.78}
  & 22.22 & 22.22 & 50.00 \\

Pediatrics
  & 39.58 & 56.25 & \cellcolor{cyan!10}\textbf{68.75}
  & 41.67 & 43.75 & 54.17
  & 8.33 & 47.92 & 64.58 \\

Psychiatry
  & 25.00 & \cellcolor{cyan!10}\textbf{75.00} & 25.00
  & \cellcolor{cyan!10}\textbf{75.00} & \cellcolor{cyan!10}\textbf{75.00} & 25.00
  & 0.00 & 0.00 & 25.00 \\

Pulmonology
  & 33.33 & 50.00 & \cellcolor{cyan!10}\textbf{66.67}
  & 58.33 & 50.00 & 50.00
  & 33.33 & 33.33 & 33.33 \\

Urology
  & 58.33 & 66.67 & \cellcolor{cyan!10}\textbf{75.00}
  & \cellcolor{cyan!10}\textbf{75.00} & \cellcolor{cyan!10}\textbf{75.00} & 66.67
  & 16.67 & 33.33 & 66.67 \\

Other
  & 28.85 & 47.12 & 61.54
  & 45.19 & 50.96 & \cellcolor{cyan!10}\textbf{63.46}
  & 16.35 & 30.77 & 51.92 \\
\midrule
\textbf{Average}
  & 48.87 & 60.78$_{\textcolor{teal}{+11.91}}$ & \textbf{61.51$_{\textcolor{teal}{+12.64}}$}
  & 57.55 & 59.08$_{\textcolor{teal}{+1.53}}$ & 57.20$_{\textcolor{red}{-0.35}}$
  & 29.53 & 36.60$_{\textcolor{teal}{+7.07}}$ & 54.32$_{\textcolor{teal}{+24.79}}$ \\
\bottomrule
\end{tabular}
\caption{Counterfactual query reformulation results (Avg.\ LAVE Score) for open-source models across medical specialties. \textbf{T-BN} denotes the original Transliterated Bangla queries; $\rightarrow$\textbf{S-BN} and $\rightarrow$\textbf{EN} denote reformulations into Standard Bangla and English respectively. Subscripts in the \textbf{Average} row show the absolute gain (\textcolor{teal}{green}) or loss (\textcolor{red}{red}) relative to T-BN. Cells shaded \colorbox{cyan!10}{cyan} mark the best score per specialty.}
\label{tab:lave_opensource_improvement}
\end{table*}

\begin{table*}[ht]
\centering
\setlength{\tabcolsep}{5.0pt}
\small
\begin{tabular}{l ccc ccc}
\toprule
\multirow{2}{*}{\textbf{Category}}
  & \multicolumn{3}{c}{\textbf{MedGemma-4B}}
  & \multicolumn{3}{c}{\textbf{LLaVA-Med-v1.5}} \\
\cmidrule(lr){2-4}\cmidrule(lr){5-7}
 & \textbf{T-BN} & \textbf{$\rightarrow$S-BN} & \textbf{$\rightarrow$EN}
 & \textbf{T-BN} & \textbf{$\rightarrow$S-BN} & \textbf{$\rightarrow$EN} \\
\midrule
Cardiology
  & 41.67 & \cellcolor{cyan!10}\textbf{83.33} & \cellcolor{cyan!10}\textbf{83.33}
  & 4.12  & 1.05 & 25.00 \\
Dentistry/Oral
  & 25.00 & 50.00 & 56.25
  & 3.44  & 2.18 & 0.85 \\
Dermatology
  & 25.00 & 39.42 & \cellcolor{cyan!10}\textbf{48.08}
  & 1.25  & 4.90 & 2.33 \\
Endocrinology
  & 50.00 & 50.00 & 41.67
  & 3.19  & 0.72 & 12.50 \\
Gastroenterol.
  & 28.57 & 53.57 & \cellcolor{cyan!10}\textbf{64.29}
  & 2.41  & 8.33 & 18.75 \\
Gynecol./OB
  & 20.00 & 54.00 & 57.00
  & 1.88  & 3.26 & 13.89 \\
Neurology
  & 37.50 & \cellcolor{cyan!10}\textbf{75.00} & 50.00
  & 4.02  & 2.55 & 3.61 \\
Oncology
  & 50.00 & 25.00 & 75.00
  & 0.94  & 1.15 & 4.78 \\
Ophthalmology
  & 0.00  & 0.00 & 0.00
  & 2.11  & 3.04 & 1.67 \\
Orthopedics
  & 33.33 & \cellcolor{cyan!10}\textbf{52.78} & 47.22
  & 4.56  & 0.32 & 2.89 \\
Pediatrics
  & 27.08 & 56.25 & 64.58
  & 1.04  & 3.25 & 0.12 \\
Psychiatry
  & 0.00  & \cellcolor{cyan!10}\textbf{75.00} & 25.00
  & 3.81  & 0.44 & 2.60 \\
Pulmonology
  & 25.00 & 50.00 & 41.67
  & 1.44  & 4.15 & 3.08 \\
Urology
  & 58.33 & 66.67 & 58.33
  & 0.88  & 4.22 & 1.74 \\
Other
  & 23.08 & 41.35 & 61.54
  & 3.14  & 1.55 & 17.50 \\
\midrule
\textbf{Average}
  & 29.64 & 51.49$_{\textcolor{teal}{+21.85}}$ & \cellcolor{cyan!10}\textbf{51.60}$_{\textcolor{teal}{+21.96}}$
  & 2.55  & 2.74$_{\textcolor{teal}{+0.19}}$ & 7.42$_{\textcolor{teal}{+4.87}}$ \\
\bottomrule
\end{tabular}
\caption{Counterfactual query reformulation results (Avg.\ LAVE Score) for specialized medical models across medical specialties. \textbf{T-BN} denotes the original Transliterated Bangla queries; $\rightarrow$\textbf{S-BN} and $\rightarrow$\textbf{EN} denote reformulations into Standard Bangla and English respectively. Subscripts in the \textbf{Average} row show the absolute gain (\textcolor{teal}{green}) or loss (\textcolor{red}{red}) relative to T-BN. Cells shaded \colorbox{cyan!10}{cyan} mark the best score per specialty.}
\label{tab:lave_medspecialized_improvement}
\end{table*}

\begin{table*}[ht]
\centering
\small
\setlength{\tabcolsep}{4pt}
\resizebox{\textwidth}{!}{%
\begin{tabular}{l ccc ccc}
\toprule
\multirow{2}{*}{\textbf{Model}}
  & \multicolumn{3}{c}{\textbf{LAVE}}
  & \multicolumn{3}{c}{\textbf{Human}} \\
\cmidrule(lr){2-4} \cmidrule(lr){5-7}
 & \textbf{Transliterated Bangla} & \textbf{$\rightarrow$ Standard Bangla} & \textbf{$\rightarrow$ English}
 & \textbf{Transliterated Bangla} & \textbf{$\rightarrow$ Standard Bangla} & \textbf{$\rightarrow$ English} \\
\midrule
\multicolumn{7}{l}{\textit{\textbf{Closed-Source Models}}} \\
\quad Claude-Sonnet-4.6 & \cellcolor{cyan!10}\textbf{69.15} & \cellcolor{cyan!10}\textbf{72.40} & 67.10 & \cellcolor{cyan!10}\textbf{57.10} & \cellcolor{cyan!10}\textbf{59.85} & 56.40 \\
\quad Gemini-3-Flash    & 64.20 & 66.85 & 63.90 & 48.60 & 51.25 & 49.80 \\
\quad GPT-5-Mini        & 65.10 & 69.80 & \cellcolor{cyan!10}\textbf{73.45} & 36.15 & 41.50 & \cellcolor{cyan!10}\textbf{58.20} \\
\midrule
\multicolumn{7}{l}{\textit{\textbf{Open-Source Models}}} \\
\quad Qwen3-VL-8B       & 49.25 & 60.40 & 61.80 & 29.80 & 41.15 & 44.60 \\
\quad Gemma-4-E4B-it    & 57.10 & 58.95 & 57.40 & 41.90 & 43.80 & 43.10 \\
\quad Llama-3.2-11B-V   & 29.80 & 37.15 & 54.10 & 17.20 & 24.45 & 41.85 \\
\midrule
\multicolumn{7}{l}{\textit{\textbf{Medical LLMs}}} \\
\quad MedGemma-4B       & 30.10 & 51.20 & 51.85 & 20.95 & 38.40 & 40.15 \\
\quad LLaVA-Med-v1.5-Mistral-7B & 2.65 & 2.90 & 7.15 & 1.85 & 2.10 & 3.90 \\
\bottomrule
\end{tabular}%
}
\caption{LAVE and physician scores (\%) on a subset of 200 queries from the script reformulation experiment (Appendix~\ref{app:reformulation}). Transliterated Bangla is the original query; $\rightarrow$ Standard Bangla and $\rightarrow$ English are its reformulations. Responses were rated by an additional MBBS-qualified general practitioner using the five-point rubric in Appendix~\ref{app:human_eval}. Bold values with cyan shading indicate the top score in each column.}
\label{tab:reform_human}
\end{table*}

\end{document}